\documentclass[final]{cvpr}

\usepackage{times}
\usepackage{epsfig}
\usepackage{graphicx}
\usepackage{amsmath}
\usepackage{amssymb}
\usepackage{subfigure}
\usepackage{color}
\usepackage{multirow}
\usepackage{diagbox}
\usepackage{array}
\usepackage{booktabs}
\usepackage{overpic}
\usepackage[table]{xcolor}
\makeindex

\graphicspath{{./figs/}}

\def\ie{\emph{i.e.~}}
\def\eg{\emph{e.g.,~}}

\usepackage{pifont}
\usepackage{overpic}
\newcommand{\cmark}{\ding{51}}%
\newcommand{\xmark}{\ding{55}}%

\newcommand{\myPara}[1]{\vspace{.05in}\noindent\textbf{#1.}}

\usepackage{url}

\newcommand{\newparam}[1]{\vspace{5pt}\noindent\textbf{#1}}%

\newcommand{\addFig}[1]{}
\newcommand{\addFigs}[1]{}

\definecolor{mygreen}{RGB}{0,100,0}
\definecolor{myred}{RGB}{150,0,0}
\usepackage[pagebackref=false,colorlinks=true,linkcolor=myred,citecolor=mygreen,bookmarks=false]{hyperref}

\def\cvprPaperID{759} 
\def\confYear{CVPR 2020}

\begin{document}

\title{Strip Pooling for Scene Parsing Networks}
\title{Strip Pooling: Rethinking Spatial Pooling for Scene Parsing}

\author{Qibin Hou$^1$ \qquad Li Zhang$^2$ \qquad Ming-Ming Cheng$^3$ \qquad Jiashi Feng$^1$ \\
  $^1$National University of Singapore \qquad $^2$University of Oxford \qquad $^3$CS, Nankai University\\
}

\maketitle


\begin{abstract}
Spatial pooling has been proven highly effective in capturing long-range contextual information for pixel-wise prediction tasks, such as scene parsing.
In this paper, beyond conventional spatial pooling that usually has a regular shape of $N\times N$,
we rethink the formulation of spatial pooling by introducing a new pooling strategy, called strip pooling,
which considers a long but narrow kernel, i.e., $1 \times N$ or $N \times 1$.
Based on strip pooling, we further investigate spatial pooling architecture design by 
1) introducing a new strip pooling module that enables backbone networks to efficiently model long-range dependencies,
2) presenting a novel building block with diverse spatial pooling as a core, and 
3) systematically comparing the performance of the proposed strip pooling and conventional spatial pooling techniques.
Both novel pooling-based designs are lightweight and can serve as an efficient plug-and-play module in existing scene parsing networks.
Extensive experiments on popular benchmarks (e.g., ADE20K and Cityscapes)
demonstrate that our simple approach establishes new state-of-the-art results.
Code is available at \url{https://github.com/Andrew-Qibin/SPNet}.
\end{abstract}

\section{Introduction} \label{sec:introduction}

Scene parsing, also known as semantic segmentation, aims to assign a semantic label to each pixel in an image.
As one of the most fundamental tasks, it has been applied in a wide range of computer vision and graphics applications~\cite{ChengSurveyVM2017}, 
such as autonomous driving~\cite{teichmann2018multinet}, medical diagnosis~\cite{ronneberger2015u}, image/video editing~\cite{murali2019single,le2019object}, salient object detection~\cite{BorjiCVM2019}, and aerial image analysis~\cite{maggiori2017high}.
Recently, methods~\cite{long2015fully,chen2017deeplab} based on fully convolutional networks (FCNs) have made extraordinary progress in scene parsing with their ability to capture  high-level semantics.
However, these approaches mostly stack \textit{local} convolutional and pooling operations, thus are hardly able to well cope with complex scenes with a variety of different categories due to the limited effective fields-of-view~\cite{zhao2016pyramid,huang2018ccnet}.

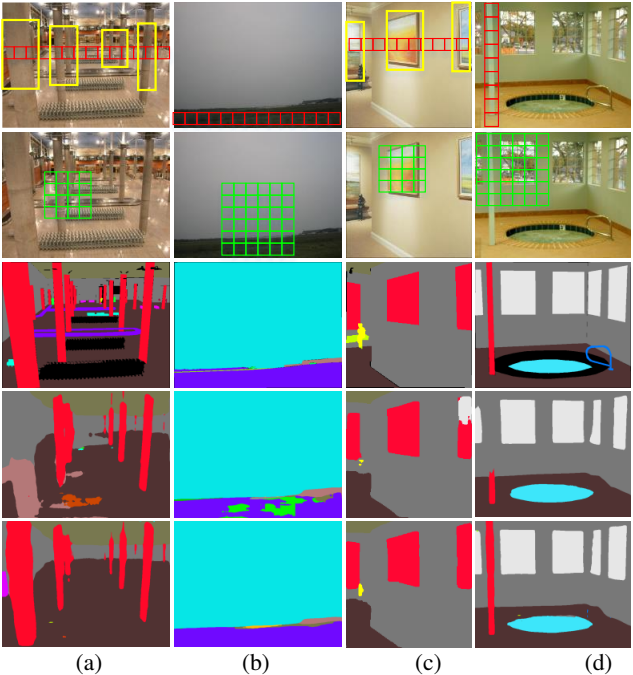
\begin{figure}
  \centering
  \small
  \begin{overpic}[width=0.48\textwidth]{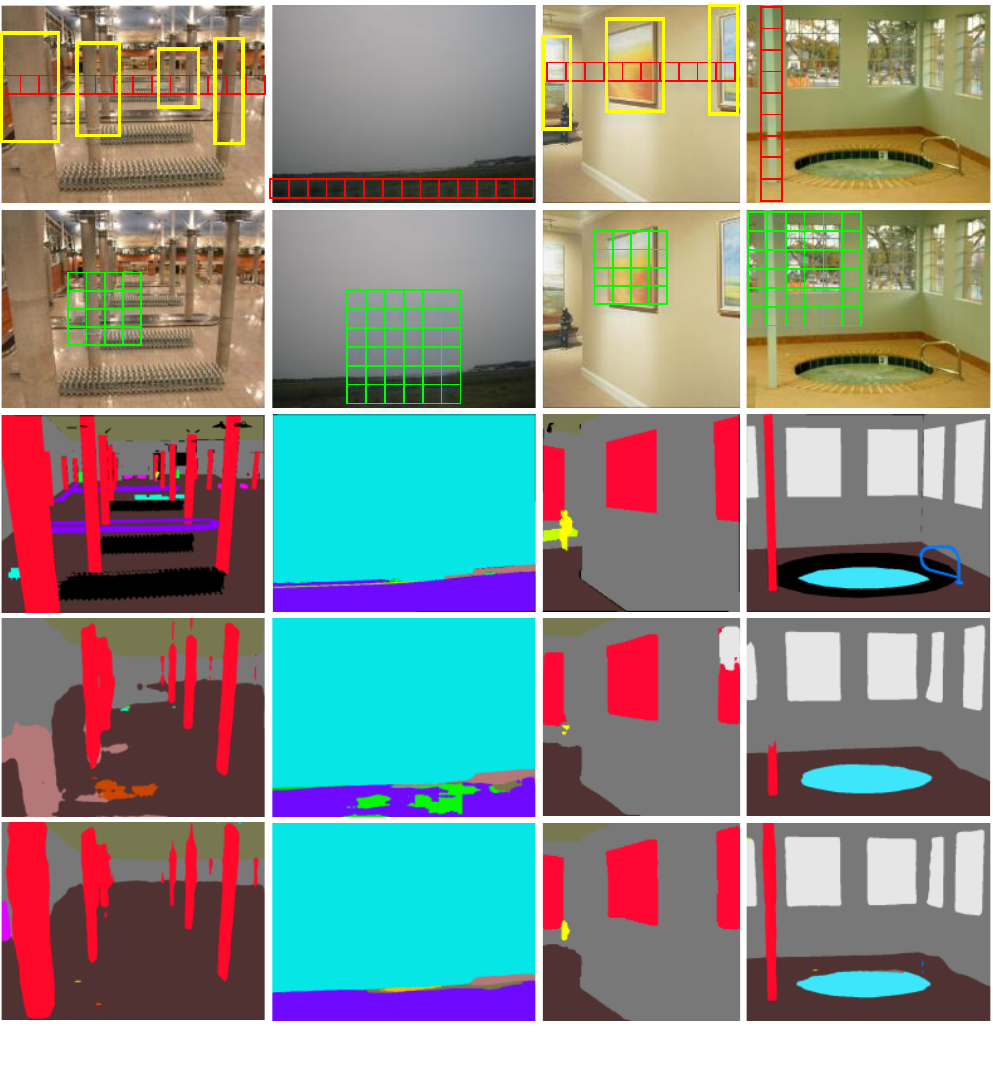}
    \put(11, 1){(a)}
    \put(35.5,  1){(b)}
    \put(61,  1){(c)}
    \put(86,  1){(d)}
    \end{overpic}
  \caption{Illustrations on how strip pooling
  and spatial pooling work differently for scene parsing.
  From top to bottom: strip pooling; conventional spatial pooling;
  ground-truth annotations; our results with conventional 
  spatial pooling only; our results with
  strip pooling considered. As shown in the top row, compared to conventional
  spatial pooling (green grids), strip pooling has a kernel of 
  band shape (red grids) and hence can capture long-range
  dependencies between regions distributed discretely
  (yellow bounding boxes).
  }
  \label{fig:scenes}
\end{figure}

One way to improve the capability of modeling the long-range dependencies in CNNs is to adopt self-attention or non-local modules~\cite{wang2018non,huang2018ccnet,chen2016attention,ren2017end,hong2016learning,yang2014context,zhao2018psanet,zhang2020dynamic,zhang2019dual,li2019global}.
However, they notoriously consume huge memory for computing the large affinity matrix at each spatial position.
Other methods for long-range context modeling include:
dilated convolutions~\cite{chen2017deeplab,chen2018encoder,chen2017rethinking,yu2015multi} 
that aim to widen the receptive fields of CNNs
without introducing extra parameters;
or global/pyramid pooling \cite{lazebnik2006beyond,zhao2016pyramid,he2019adaptive,chen2017deeplab,chen2018encoder,yang2018denseaspp} that summarizes global clues of the images.
However, a common limitation for these 
methods, including dilated convolutions and pooling, 
is that they all probe the input features map within square windows.
This limits their flexibility in capturing anisotropy context that widely exists in realistic scenes. 
For instance, in some cases, the target objects
may have long-range banded structure (\eg the grassland in Figure~\ref{fig:scenes}b) or distributed discretely (\eg the pillars in Figure~\ref{fig:scenes}a).
Using large square pooling windows cannot well solve the problem because it would inevitably 
incorporate contaminating information from irrelevant regions \cite{he2019adaptive}.

In this paper, to more efficiently and effectively capture long-range dependencies, we exploit spatial pooling for enlarging the receptive fields of CNNs and collecting informative contexts, and present the concept 
of \emph{strip pooling}.
As an alternative to global pooling, strip pooling offers two advantages.
First, it deploys a long kernel shape along one spatial dimension and hence enables capturing long-range relations of isolated regions, as shown in the top part of Figures~\ref{fig:scenes}a and \ref{fig:scenes}c.
Second, it keeps a narrow kernel shape along 
the other spatial dimension, which facilitates capturing local context and prevents irrelevant regions from interfering the label prediction.
Integrating such long but narrow pooling kernels enables the scene parsing networks to  simultaneously aggregate both global and local context. 
This is essentially different from the traditional spatial pooling which collects context from a fixed square region.

Based on the strip pooling operation, we present two pooling based modules for scene parsing networks.
First, we design a \textit{Strip Pooling Module} (SPM) to effectively enlarge the receptive field of the backbone.
More concretely, the SPM consists of two pathways, which focus on encoding long-range context along either the horizontal or vertical spatial dimension.
For each spatial location in the pooled map, it encodes its globally horizontal and vertical information and then uses the encodings to balance its own weight for feature refinement.
Furthermore, we present a novel add-on
residual building block, called the \textit{Mixed Pooling module} (MPM), to further model long-range dependencies at high semantic level.
It gathers informative contextual information by exploiting pooling operations with different kernel shapes to probe the images with complex scenes.
To demonstrate the effectiveness of the proposed
pooling-based modules, we present SPNet which incorporates both modules into the ResNet \cite{He2016} backbone.
Experiments show that our SPNet establishes new state-of-the-art results on popular scene parsing benchmarks.

The contributions of this work are as follows:
(i) 
We investigate the conventional design of the spatial pooling and present the concept of \emph{strip pooling}, which 
inherits the merits of global average pooling to collect long-range dependencies and meanwhile focus on local details.
(ii) 
We design a \emph{Strip Pooling Module} and a \emph{Mixed Pooling Module} based on strip pooling. 
Both modules are lightweight and can serve as efficient add-on blocks to be plugged into any 
backbone networks to generate high-quality segmentation predictions.
(iii) 
We present SPNet integrating the above two pooling-based
modules into a single architecture, which achieves 
significant improvements over the baselines and 
establishes new state-of-the-art results on widely-used
scene parsing benchmark datasets.

\section{Related Work} \label{sec:related_work}

Current state-of-the-art scene parsing (or semantic segmentation) 
methods mostly leverage convolutional neural networks (CNNs).
However, the receptive fields of CNNs grow slowly by stacking the local convolutional or pooling operators, which therefore hampers them from taking enough useful contextual information into account.
Early techniques for modeling contextual relationships for scene parsing involve the conditional random fields (CRFs)~\cite{krahenbuhl2012efficient,vemulapalli2016gaussian,arnab2016higher,zheng2015conditional}.
They are mostly modeled in the discrete label space and computationally expensive, thus are now less successful for producing state-of-the-art results of scene parsing albeit have been integrated into CNNs.

For continuous feature space learning, prior work use multi-scale feature aggregation \cite{long2015fully,chen2017deeplab,lin2016efficient,hariharan2015hypercolumns,noh2015learning,lin2018multi,lin2017refinenet,badrinarayanan2017segnet,peng2017large,bulo2017loss,tian2019decoders,pami20Res2net} to fuse the contextual information by probing the incoming features with filters or pooling operations at multiple rates and multiple fields-of-view.
DeepLab~\cite{chen2017deeplab,chen2017rethinking} and its follow-ups~\cite{chen2018encoder,yang2018denseaspp,mehta2018espnet} adopt dilated convolutions and fuse different dilation rate features to increase the receptive filed of the network.
Besides, aggregating non-local context \cite{liu2017learning,yuan2018ocnet,li2019expectation,ding2018context,chen2016attention,ren2017end,hong2016learning,yang2014context,zhao2018psanet,huang2018ccnet,ding2019semantic} 
is also effective for scene parsing.

Another line of research on improving the receptive field is the spatial pyramid pooling~\cite{zhao2016pyramid,he2019adaptive}. 
By adopting a set of parallel pooling operations with a unique kernel size at each pyramid level, the network is able to capture large-range context.
It has been shown promising on several scene parsing benchmarks.
However, its ability to exploit contextual information is limited since only square kernel shapes are applied.
Moreover, the spatial pyramid pooling is only modularized on top of the backbone network thus rendering it is not flexible or 
directly applicable in the network building block for feature learning.
In contrast, our proposed \textit{strip pooling} module and \textit{mixed pooling} module adopt pooling kernels with size $1 \times N$ or $N \times 1$, both of which can be plugged and stacked into existing networks.
This difference enables the network to exploit 
rich contextual relationships in each of the proposed building blocks.
The proposed modules have proven to be much more powerful and adaptable than the spatial pyramid pooling in our experiments.

\section{Methodology} \label{sec:method}

\begin{figure*}
  \centering
  \includegraphics[width=0.9\linewidth]{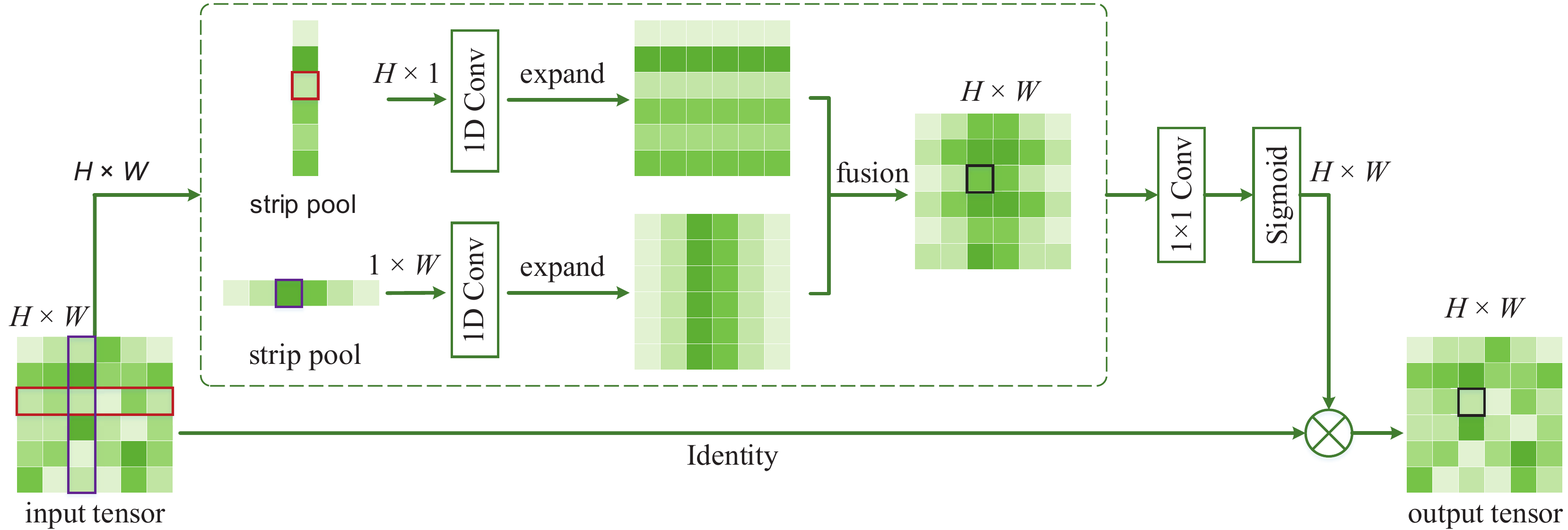}
  \caption{
  Schematic illustration of the Strip Pooling (SP) module.
  }
  \label{fig:strip_pool}
\end{figure*}
In this section, we first give the concept of \emph{strip pooling} and then introduce two model designs based on strip pooling to demonstrate how it improves scene parsing networks.
Finally, we describe the entire architecture of the proposed scene parsing network augmented by strip pooling.

\subsection{Strip Pooling} \label{sec:strip_pooling}

Before describing the formulation of strip pooling, we first briefly review the average pooling operation.

\newparam{Standard Spatial Average Pooling:}
Let $\mathbf{x} \in \mathbb{R}^{H \times W}$ be a two-dimensional input tensor, where $H$ and $W$ are the spatial height and width, respectively.
In an average pooling layer, a spatial extent of the pooling ($h \times w$) is required.
Consider a simple case where $h$ divides $H$ and $w$ divides $W$. Then 
the output $\mathbf{y}$ after pooling is also a two-dimensional tensor with height $H_{o} = \frac{H}{h}$ and width $W_{o} = \frac{W}{w}$.
Formally, the average pooling operation can be 
written as 
\begin{equation} \label{eqn:avg_pool}
y_{i_{o}, j_{o}} = \frac{1}{h \times w} \sum_{0 \le i < h}\sum_{0 \le j < w}x_{i_{o} \times h + i, j_{o} \times w + j},
\end{equation}
where $0 \le i_{o} < H_{o}$ and $0 \le j_{o} < W_{o}$.
In Eqn.~\ref{eqn:avg_pool}, each spatial location of $\mathbf{y}$ corresponds to a pooling window of size $h \times w$.
The above pooling operation has been successfully applied to previous work \cite{zhao2016pyramid,he2019adaptive} for collecting long-range context.
However, it may unavoidably incorporate lots of irrelevant regions when processing objects with irregular shapes as shown in Figure~\ref{fig:scenes}.

\newparam{Strip Pooling:}
To alleviate the above problem, we present the concept of `strip pooling' here, which uses a band shape pooling window to perform pooling along either the horizontal or the vertical dimension, as shown in the top row of Figure~\ref{fig:scenes}.
Mathematically, given the two-dimensional tensor $\mathbf{x} \in \mathbb{R}^{H \times W}$, in strip pooling, a spatial extent of pooling $(H, 1)$ or $(1, W)$ is required.
Unlike the two-dimensional average pooling, the proposed strip pooling averages all the feature values in a row or a column. 
Thus, the output $\mathbf{y}^h \in \mathbb{R}^{H}$ after horizontal strip pooling can be written as
\begin{equation} \label{eqn:strip_pool_h}
y_{i}^h = \frac{1}{W} \sum_{0 \le j < W}x_{i, j}.
\end{equation}
Similarly, the output $\mathbf{y}^v \in \mathbb{R}^{W}$ after vertical strip pooling can be written as
\begin{equation} \label{eqn:strip_pool_v}
y_{j}^v = \frac{1}{H} \sum_{0 \le i < H}x_{i, j}.
\end{equation}
Given the horizontal and vertical strip pooling layers, it is easy to build long-range dependencies between regions distributed discretely and encode regions with the banded shape, thanks to the long and narrow kernel shape. 
Meanwhile, it also focuses on capturing local details due to its narrow kernel shape along 
the other dimension.
These properties make the proposed strip pooling different from conventional spatial pooling that relies on square-shape kernels.
In the following, we will describe how to leverage strip pooling (Eqn.~\ref{eqn:strip_pool_h} and Eqn.~\ref{eqn:strip_pool_v}) to improve scene parsing networks.

\subsection{Strip Pooling Module} \label{sec:sp_block}

It has been demonstrated in previous work \cite{chen2018encoder,fu2019dual} that enlarging the receptive fields of the backbone networks is beneficial to scene parsing.
In this subsection, motivated by this fact,
we introduce an effective way to help backbone networks capture long-range context by exploiting strip pooling.
In particular, we present a novel \emph{Strip Pooling} module (SPM), which leverages both horizontal and vertical strip pooling operations to gather long-range context from different
spatial dimensions.
Figure~\ref{fig:strip_pool} depicts our proposed
SPM.
Let $\mathbf{x} \in \mathbb{R}^{C \times H \times W}$ be an input tensor, where $C$ denotes the number of channels.
We first feed $\mathbf{x}$ into two parallel pathways, each of which contains a horizontal or vertical strip pooling layer 
followed by a 1D convolutional layer with kernel size 3 for modulating the current location and its neighbor features. 
This gives $\mathbf{y}^h \in \mathbb{R}^{C \times H}$ and $\mathbf{y}^v \in \mathbb{R}^{C \times W}$.
To obtain an output $\mathbf{z} \in \mathbb{R}^{C \times H \times W}$ that contains more useful global priors, we first combine $\mathbf{y}^h$  and $\mathbf{y}^w$ together as follows, yielding 
$\mathbf{y} \in \mathbb{R}^{C \times H \times W}$:
\begin{equation}
    y_{c,i,j} = y^h_{c,i} + y^v_{c,j}.
\end{equation}
Then, the output $\mathbf{z}$ is computed as
\begin{equation} \label{eq:reweight}
\mathbf{z} = \mbox{Scale}(\mathbf{x}, ~\sigma(f(\mathbf{y}))),
\end{equation}
where $\mbox{Scale}(\cdot, ~\cdot)$ refers to element-wise multiplication, $\sigma$ is the sigmoid function and $f$ is a $1\times1$ convolution.
It should be noted that there are multiple ways to combine the features extracted by the two strip pooling layers, such as computing the inner product between two extracted 1D feature vectors.
However, taking the efficiency into account and to make the SPM lightweight, we adopt the operations described above, which we find still work well.

In the above process, each position in the output tensor is allowed to build relationships with a variety of positions in the input tensor.
For example, in Figure~\ref{fig:strip_pool}, the square bounded by the black box in the output tensor is connected to all the locations with the same horizontal or vertical coordinate as it
(enclosed by red and purple boxes).
Therefore, by repeating the above aggregation process a couple of times, it is possible to build long-range dependencies over the whole scene.
Moreover, benefiting from the element-wise multiplication 
operation, the proposed SPM can also be considered as an 
attention mechanism and directly applied to any pretrained 
backbone networks \emph{without training them
from scratch}.

Compared to global average pooling, strip pooling considers long but narrow ranges instead of the whole feature map, 
avoiding most unnecessary connections to be built between locations that are far from each other.
Compared to attention-based modules \cite{fu2019dual,he2019adaptive}
that need a large amount of computation to build relationships between each pair of locations, our SPM is lightweight and can be easily embedded into any building blocks to improve the capability of capturing long-range spatial dependencies and exploiting inter-channel dependencies.
We will provide more analysis on the performance of our approach against existing attention-based methods.

\subsection{Mixed Pooling Module} \label{sec:mpmodule}

It turns out that the pyramid pooling module (PPM) 
is an effective way to enhance scene parsing networks \cite{zhao2016pyramid}.
However, PPM heavily relies on the standard spatial
pooling operations (albeit with different pooling kernels at different pyramid levels), making it still suffers as analyzed in Section~\ref{sec:strip_pooling}.
Taking into account the advantages of both standard
spatial pooling and the proposed strip pooling, 
we advance the PPM and design a Mixed Pooling
Module (MPM) which focuses on aggregating different 
types of contextual information via various pooling 
operations to make the feature representations
more discriminative.

\begin{figure}
  \centering
  \includegraphics[width=\linewidth]{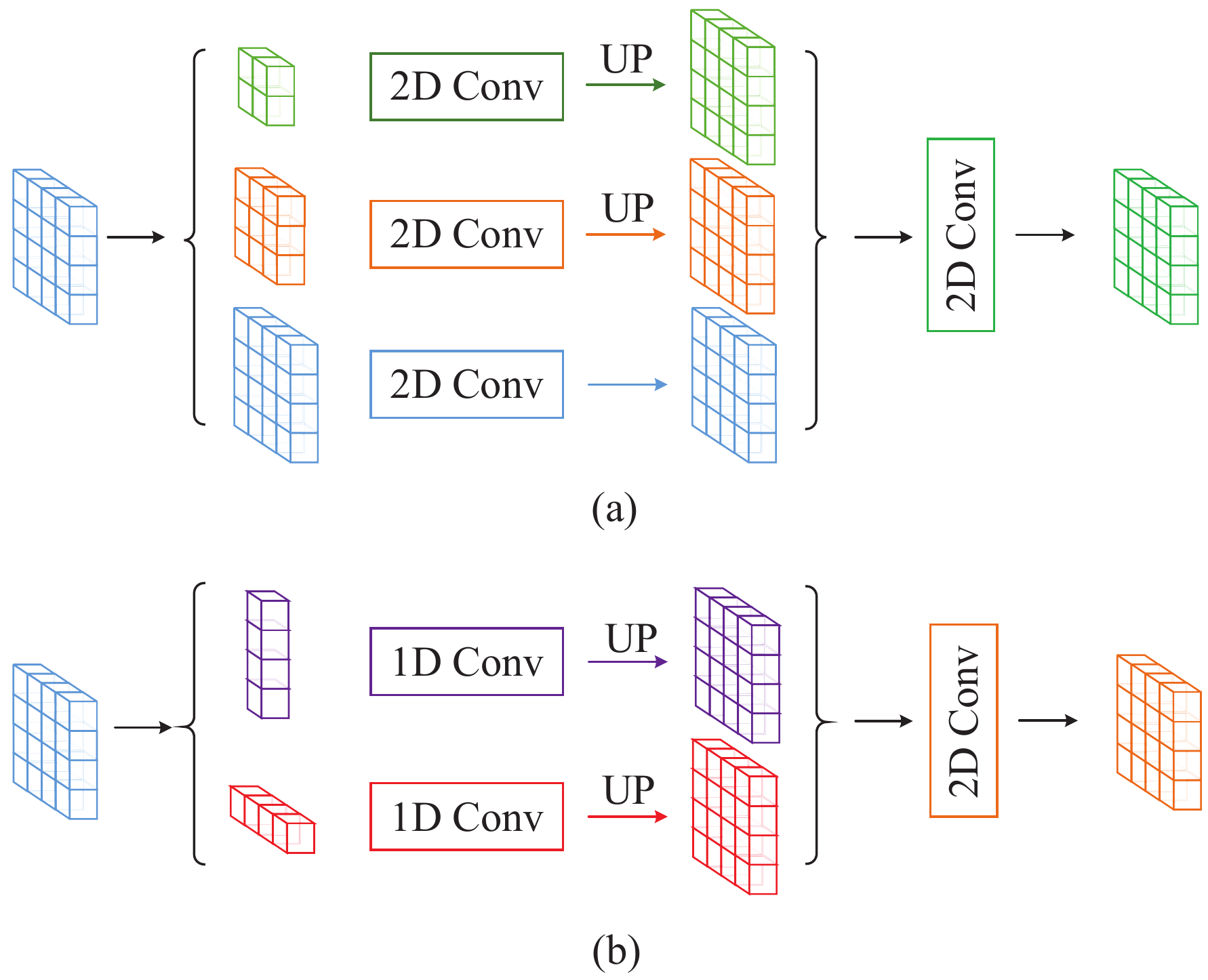}
  \caption{(a) Short-range dependency aggregation sub-module. 
  (b) Long-range dependency aggregation sub-module. 
  Inspired by \cite{lin2017feature,liu2019simple}, a convolutional layer is added after the fusion operation in each sub-module to reduce the aliasing effect brought by down-sampling operations.
  }
  \label{fig:network}
\end{figure}

The proposed MPM consists of two sub-modules that 
simultaneously capture short-range and long-range
dependencies among different locations,
which we find are both essential for scene parsing networks.
For long-range dependencies, unlike previous work
\cite{zhang2018context,zhao2016pyramid,chen2018encoder}
that use the global average pooling layer, we propose to
gather such kind of clues by employing both horizontal
and vertical strip pooling operations.
A simplified diagram can be found in 
Figure~\ref{fig:network}(b).
As analyzed in Section~\ref{sec:sp_block}, the strip 
pooling makes connections among regions distributed 
discretely over the whole scene 
and encoding regions with banded structures possible.
However, for cases where semantic regions are
distributed closely, spatial pooling is also 
necessary for capturing local contextual information.
Taking this into account, as depicted
in Figure~\ref{fig:network}(a),
we adopt a lightweight pyramid pooling sub-module
for short-range dependency collection.
It has two spatial pooling layers followed by 
convolutional layers for multi-scale feature extraction 
plus a 2D convolutional layer 
for original spatial information preserving.
The feature maps after each pooling are with bin sizes 
of $20 \times 20$ and $12 \times 12$, respectively.
All three sub-paths are then combined by summation.

Based on the above two sub-modules, we propose to nest them into residual blocks \cite{He2016} with bottleneck structure for parameter reduction and modular design.
Specifically, before each sub-module, an $1\times1$ convolutional layer is first used for channel reduction.
The outputs from both sub-modules are concatenated together and then fed into another $1\times1$ convolutional layer for channel expansion as done in \cite{He2016}.
Note that all convolutional layers, aside from the ones for channel reduction and expansion, are with kernel size $3\times3$ or 3 (for 1D convolutional layers).

It is worth mentioning that unlike the spatial pyramid pooling modules~\cite{zhao2016pyramid,chen2018encoder}, the proposed MPM is a kind of modularized design.
The advantage is that it can be easily used in a sequential way to expand the role of the long-range dependency collection sub-module.
We find that with the same backbone our network with only two MPMs (around 1/3 parameters of the original PPM \cite{zhao2016pyramid}) performs even better than the PSPNet.
In our experiment section, we will provide more  results and analysis on this.

\subsection{Overall Architecture} \label{sec:architecture}

Based on the proposed SPM and MPM, we introduce
an overall architecture, called SPNet,
in this subsection.
We adopt the classic residual networks \cite{He2016} as our backbones.
Following \cite{chen2017deeplab,zhao2016pyramid,fu2019dual}, we improve the original ResNet 
with the dilation strategy and the final feature map size is set to 1/8 of the input image.
The SPMs are added after the $3\times3$ convolutional layer of the last building block in each stage and all building blocks in the last stage.
All convolutional layers in an SPM share the same number of channels to the input tensor.

For the MPM, we directly build it upon the
backbone network because of its modular design.
Since the output of the backbone is with 2048 channels, we first connect a $1\times1$ convolutional layer to the backbone to reduce the output channels from 2048 to 1024 and then add two MPMs.
In each MPM, following \cite{He2016}, all 
convolutional layers with kernel size $3\times3$ or 
3 have 256 channels (\ie a reduction rate of 1/4
is used).
A convolutional layer is added at the end to predict 
the segmentation map.

\section{Experiments} \label{sec:experiments}

We evaluate the proposed SPM and MPM 
on popular scene parsing datasets, including 
ADE20K~\cite{zhou2017scene},
Cityscapes~\cite{cordts2016cityscapes}, 
and Pascal Context~\cite{mottaghi2014role}. 
Moreover, we also conduct comprehensive ablation analysis on the effect of the proposed strip pooling based on the ADE20K dataset as done in \cite{zhao2016pyramid}.

\subsection{Experimental Setup}

Our network is implemented based on two public toolboxes~\cite{semseg2019,encoding2018} 
and Pytorch~\cite{paszke2019pytorch}.
We use 4 GPUs to run all the experiments.
The batch size is set to 8 for Cityscapes and 16 
for other datasets during training.
Following most previous works \cite{chen2017deeplab,zhao2016pyramid,zhang2018context},
we adopt the `poly' learning rate policy (\ie the base one multiplying $(1 - \frac{iter}{max\_{iter}})^{power}$) in training.
The base learning rate is set to 0.004 for ADE20K and
Cityscapes datasets and 0.001 for the Pascal Context dataset.
The power is set to 0.9.
The training epochs are as follows: ADE20K (120), 
Cityscapes (180), and Pascal Context (100). 
Momentum and weight decay rate are set to 0.9 
and 0.0001, respectively.
We use synchronized Batch Normalization in training
as done in \cite{zhang2018context,zhao2016pyramid}.

For data augmentation, similar to \cite{zhao2016pyramid,zhang2018context},
we randomly flip and rescale the input images from 0.5 to 2 and 
finally crop the image to a fixed size of $768\times768$
for Cityscapes and $480\times480$ for others.
By default, we report results under the standard evaluation metric\textemdash mean Intersection of Union (mIoU).
For datasets with no ground-truth annotations available, we get results from
the official evaluation servers.
For all experiments, we use cross-entropy loss to optimize all models.
Following~\cite{zhao2016pyramid}, we exploit an auxiliary loss (connected to the last residual block of the forth stage) and the loss weight is set to 0.4.
We also report multi-model results to fairly compare our approach with others,
\ie averaging the segmentation probability maps from multiple image scales
$\{0.5, 0.75, 1.0, 1.25, 1.5, 1.75\}$ as 
in \cite{lin2017refinenet,zhao2016pyramid,zhang2018context}.

\begin{table}[t]
  \centering
  \small
  \setlength\tabcolsep{1.7mm}
  \renewcommand\arraystretch{1}
  \begin{tabular}{lcccc} \toprule[0.5pt]
    Settings & \#Params & SPM & mIoU & Pixel Acc \\ \midrule[0.5pt] \midrule[0.5pt]
    Base FCN     & 27.7 M   & \xmark & 37.63 & 77.60\% \\
    Base FCN + PPM \cite{zhao2016pyramid} & +21.0 M & \xmark & 41.68 & 80.04\% \\  \midrule[0.5pt] \midrule[0.5pt]
    Base FCN + 1 MPM & +4.4 M & \xmark & 40.50 & 79.60\% \\
    Base FCN + 2 MPM & +8.8 M & \xmark & 41.92 & 80.03\% \\ 
    Base FCN + 2 MPM & +11.9 M & \cmark & \textbf{44.03} & \textbf{80.65}\% \\ 
    \bottomrule[0.5pt]
  \end{tabular}
  \vspace{0pt}
  \caption{Ablation analysis on the number of mixed pooling modules (MPMs). `SPM' refers to the strip pooling module.
  As can be seen, when more MPMs are used, 
  better results are yielded. All results are based on ResNet-50 backbone and single-model test.
  Best result is highlighted in \textbf{bold}.}
  \label{tab:num_modules}
\end{table}

\subsection{ADE20K}

The ADE20K dataset \cite{zhou2017scene} is one of the most challenging benchmarks, which contains 150 classes and a variety of scenes with 1,038 image-level labels.
We follow the official protocal to split the whole dataset.
Like most previous works, we use both pixel-wise accuracy (Pixel Acc.) and mean of Intersection over Union (mIoU) for evaluation.
We also adopt multi-model test and use the averaged results for evaluation following~\cite{lin2017refinenet,zhao2016pyramid}.
For ablation experiments, we adopt ResNet-50 as our backbone as done in \cite{zhao2016pyramid}.
When comparing with prior works, we use ResNet-101.

\begin{table}[t]
  \centering
  \small
  \setlength\tabcolsep{1.0mm}
  \renewcommand\arraystretch{1.0}
  \begin{tabular}{lcccc} \toprule[0.5pt]
    Settings & w/ SPM & mIoU & Pixel Acc \\ \midrule[0.5pt] \midrule[0.5pt]
    Base FCN   & \xmark & 37.63 & 77.60\% \\
    Base FCN + 2 MPM (SRD only)  & \xmark & 40.50 & 79.34\% \\
    Base FCN + 2 MPM (LRD only) & \xmark & 41.14 & 79.64\% \\
    \midrule[0.5pt] \midrule[0.5pt]
    Base FCN + 2 MPM (SRD + LRD)  & \xmark & 41.92 & 80.03\% \\
    Base FCN + 2 MPM (SRD + LRD)  & \cmark & 44.03 & 80.65\% \\
    \bottomrule[0.5pt]
  \end{tabular}
  \vspace{0pt}
  \caption{Ablation analysis on the mixed pooling module (MPM). 
  `SPM' refers to the strip pooling module. `SRD' and `LRD' denote the short-range dependency aggregation sub-module and the long-range dependency aggregation sub-module, respectively.
  As can be seen, collecting both short-range and long-range dependencies are essential for yielding better segmentation results. 
  All results are based on single-model test.}
  \label{tab:mp_modules}
\end{table}

\subsubsection{Ablation Studies}

\newparam{Number of MPMs:}
As stated in Section~\ref{sec:mpmodule}, the MPM
is built based on the bottleneck structure of residual blocks \cite{He2016} 
and hence can be easily repeated multiple times to expand the role of strip pooling.
Here, we investigate how many MPMs are needed to balance the performance and the runtime cost of the proposed approach.
As shown in Table~\ref{tab:num_modules}, we list the results when different numbers of MPMs are used based on the ResNet-50 backbone.
One can see when no MPM is used (base FCN), 
we achieve a result of 37.63\% in terms of mIoU.
When 1 MPM is used, we have a result of 40.50\%,
i.e. around 3.0\% improvement.
Furthermore, when we add two MPMs to the backbone, 
a performance gain of around 4.3\% can be obtained.
However, adding more MPMs gives trivial performance gain.
This may be because the receptive field is already large enough. 
As a result, regarding the runtime cost, 
we set the number of MPMs to 2 by default.

To show the advantages of the proposed MPM 
over PPM \cite{zhao2016pyramid}, we also
show the result and the parameter number of 
PSPNet in Table~\ref{tab:num_modules}.
It can be easily seen that the setting of 
`Base FCN + 2 MPM' already performs better 
than PSPNet despite 12M fewer parameters than PSPNet.
This phenomenon demonstrates that our modularized design of MPM is much more effective than PPM.

\begin{figure}
    \centering 
    \small
    \begin{overpic}[width=\linewidth]{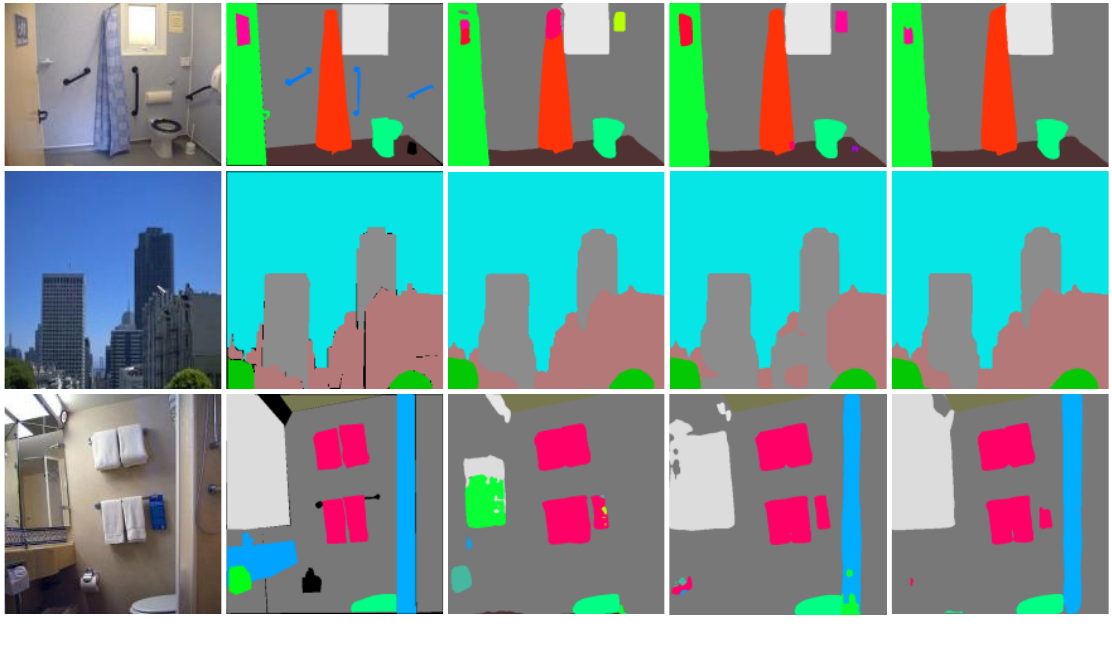}
    \put(2.5, 0.5){(a) Image}
    \put(25, 0.5){(b) GT}
    \put(42, 0.5){(c) 2 SRD}
    \put(62, 0.5){(d) 2 LRD}
    \put(81, 0.5){(e) 2 MPM}
    \end{overpic}
    \caption{Visual comparisons among different settings of the MP module
    (MPM). `2 SRD' means we use 2 MPMs with only the short-range dependency aggregation module included and `2 LRD' means we use 2 MPMs with only the long-range dependency aggregation module included. } 
    \label{fig:mp_comp} 
\end{figure}

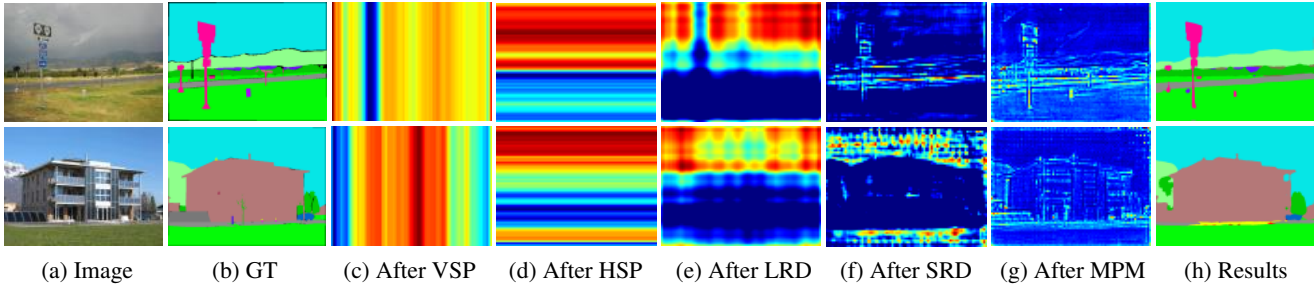
\begin{figure*}
    \centering 
    \small
    \begin{overpic}[width=\textwidth]{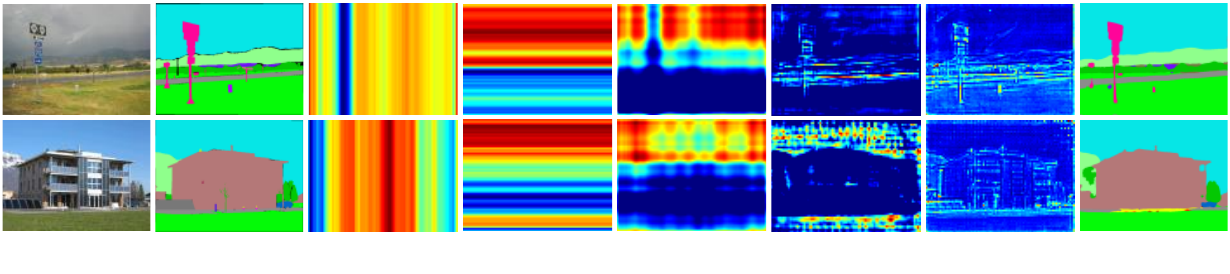}
    \put(3, 1){(a) Image}
    \put(16,  1){(b) GT}
    \put(26,  1){(c) After VSP}
    \put(38.6,  1){(d) After HSP}
    \put(51.3,  1){(e) After LRD}
    \put(63.8,  1){(f) After SRD}
    \put(76,  1){(g) After MPM}
    \put(90,  1){(h) Results}
    \end{overpic}
    \caption{Visualization of selected feature maps at different positions of the proposed MP module. \textbf{VSP}: vertical
    strip pooling; \textbf{HSP}: horizontal strip pooling;
    \textbf{SRD}: short-range dependency aggregation sub-module (Figure~\ref{fig:network}a); \textbf{LRD}: long-range dependency aggregation sub-module (Figure~\ref{fig:network}b); 
    \textbf{MPM}: mixed pooling module.} 
    \label{fig:sp_vis} 
\end{figure*}

\newparam{Effect of strip pooling in MPMs:}
It has been described in Section~\ref{sec:mpmodule} that 
the proposed MPM contains two sub-modules for 
collecting short-range and long-range dependencies, respectively.
Here, we ablate the importance of the proposed strip pooling.
The corresponding results are shown in
Table~\ref{tab:mp_modules}.
Obviously, collecting long-range dependencies with strip pooling (41.14\%) is more effective than collecting only short-range dependencies (40.5\%),
but gathering both of them further improves (41.92\%).
To further demonstrate how the strip pooling works in MPM,
we visualize some feature maps at different positions of MPM in Figure~\ref{fig:sp_vis} and some segmentation results under different settings of MPM in Figure~\ref{fig:mp_comp}.
Clearly, the proposed strip pooling can more effectively collect long-range dependencies.
For example, the feature map output from the long-range dependency aggregation module (LRD) in the top row of Figure~\ref{fig:sp_vis}
can accurately locate where the sky is.
However, global average pooling cannot do this
because it encodes the whole feature map to 
a single value.

\newparam{Effectiveness of SPMs:}
We empirically find that there is no need to add the proposed SPM to each building block of the backbone network despite its light weight.
In this experiment, we consider four scenarios, which are listed in Table~\ref{tab:sp_block}.
We take the base FCN followed by 2 MPMs as the baseline.
We first add an SPM to the last building block 
in each stage;  
the resulting mIoU score is 42.61\%.
Second, we attempt to add SPMs to all the building blocks in the last stage, and find the performance slightly declines to 42.30\%.
Next, when we add SPMs to both the above positions,
an mIoU score of 44.03\% can be yielded.
However, when we attempt to add SPMs to all the building blocks of the backbone, there is nearly no performance gain already.
Regarding the above results, by default, we add SPMs 
to the last building block of each stage and all the 
building blocks of the last stage.
In addition, when we take only the base FCN as 
our baseline and add the proposed SPMs, 
the mIoU score increases from 37.63\% to 41.66\%, 
achieving an improvement of nearly 4\%.
All the above results indicate that adding SPMs
to the backbone network does benefit the scene 
parsing networks.

\begin{table}[t]
  \centering
  \small
  \setlength\tabcolsep{1mm}
  \renewcommand\arraystretch{1.0}
  \begin{tabular}{lccccc} \toprule[0.7pt]
    Settings             & SPM Position & \#MPM & mIoU & Pixel Acc. \\ \midrule[0.5pt]\midrule[0.5pt]
    Base FCN & - & 2 & 41.92 & 80.03\% \\
    Base FCN + SPM & L & 2 & 42.61 & 80.38\% \\ 
    Base FCN + SPM & A & 2 & 42.30 & 80.22\% \\ \midrule[0.5pt]\midrule[0.5pt]
    Base FCN + SE \cite{hu2018squeeze}  & A + L & 2 & 41.34 & 80.05\% \\
    Base FCN + SPM & A + L & 0 & 41.66 & 79.69\% \\
    Base FCN + SPM & A + L & 2 & \textbf{44.03} & \textbf{80.65}\% \\ 
    \bottomrule[0.7pt]
  \end{tabular}
  \vspace{0pt}
  \caption{Ablation analysis on the strip pooling module (SPM). \textbf{L}: Last building block in each stage. \textbf{A}: All building blocks in the last stage. As can be seen, SPM can largely improve the performance of the base FCN from 37.63 to 41.66.}
  \label{tab:sp_block}
\end{table}



\begin{table}[t]
  \centering
  \small
  \setlength\tabcolsep{1.8mm}
  \renewcommand\arraystretch{1.0}
  \begin{tabular}{lcccc} \toprule[0.7pt]
    Settings & Multi-Scale + Flip & mIoU (\%) & Pixel Acc. (\%) \\ \midrule[0.5pt]\midrule[0.5pt]
    SPNet-50  &   & 44.03 & 80.65 \\
    SPNet-50 & \cmark & 45.03 & 81.32 \\ \midrule[0.5pt]\midrule[0.5pt]
    SPNet-101 &    & 44.52 & 81.37 \\
    SPNet-101 & \cmark & 45.60 & 82.09 \\
    \bottomrule[0.7pt]
  \end{tabular}
  \vspace{0pt}
  \caption{More ablation experiments when different 
  backbone networks are used.}
  \label{tab:aux_res}
\end{table}

\newparam{Strip Pooling \emph{v.s.} Global Average Pooling:} 
To demonstrate the advantages of the proposed strip pooling over the global average pooling, we attempt to change the strip pooling operations in the proposed SPM to global average pooling.
Taking the base FCN followed by 2 MPMs as
the baseline, when we add SPMs to the base FCN,
the performance increases from 41.92\% to 44.03\%.
However, when we change the proposed strip pooling 
to global average pooling as done 
in \cite{hu2018squeeze},
the performance drops from 41.92\% to 41.34\%,
which is even worse than the baseline
as shown in Table~\ref{tab:sp_block}.
This may be due to directly 
fusing feature maps to construct
a 1D vector which leads to loss of too much spatial information and hence ambiguity as pointed out in the previous work \cite{zhao2016pyramid}.

\newparam{More experiment analysis:} In this part, we
show the influence of different experiment settings on the performance, including the depth of the backbone network
and multi-scale test with flipping.
As listed in Table~\ref{tab:aux_res}, multi-scale test with flipping
can largely improve the results for both backbones.
Moreover, using deeper backbone networks also benefits the performance (ResNet-50: 45.03\% $\rightarrow$ ResNet-101: 45.60\%).

\begin{figure*}
    \centering 
    \small
    \begin{overpic}[width=\textwidth]{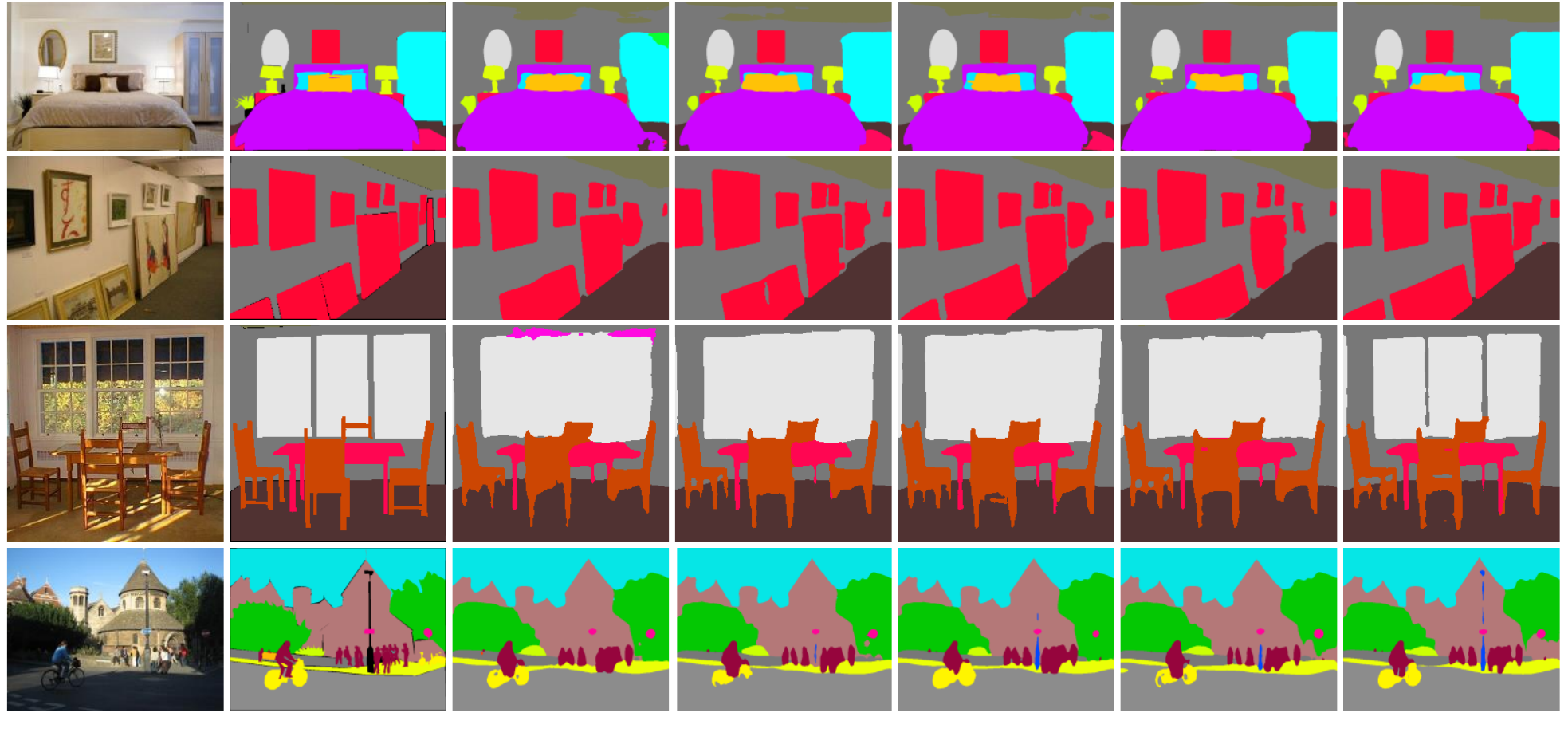}
    \put(3.5, 1){(a) Image}
    \put(19,  1){(b) GT}
    \put(31,  1){(c) Base FCN}
    \put(44.3,  1){(d) 1 MPM only}
    \put(58.5,  1){(e) 2 MPM only}
    \put(74,  1){(f) SPM only}
    \put(89,  1){(g) SPNet}
    \end{overpic}
    \caption{Visual results of the proposed approach under different model settings.} 
    \label{fig:vis_res} 
\end{figure*}
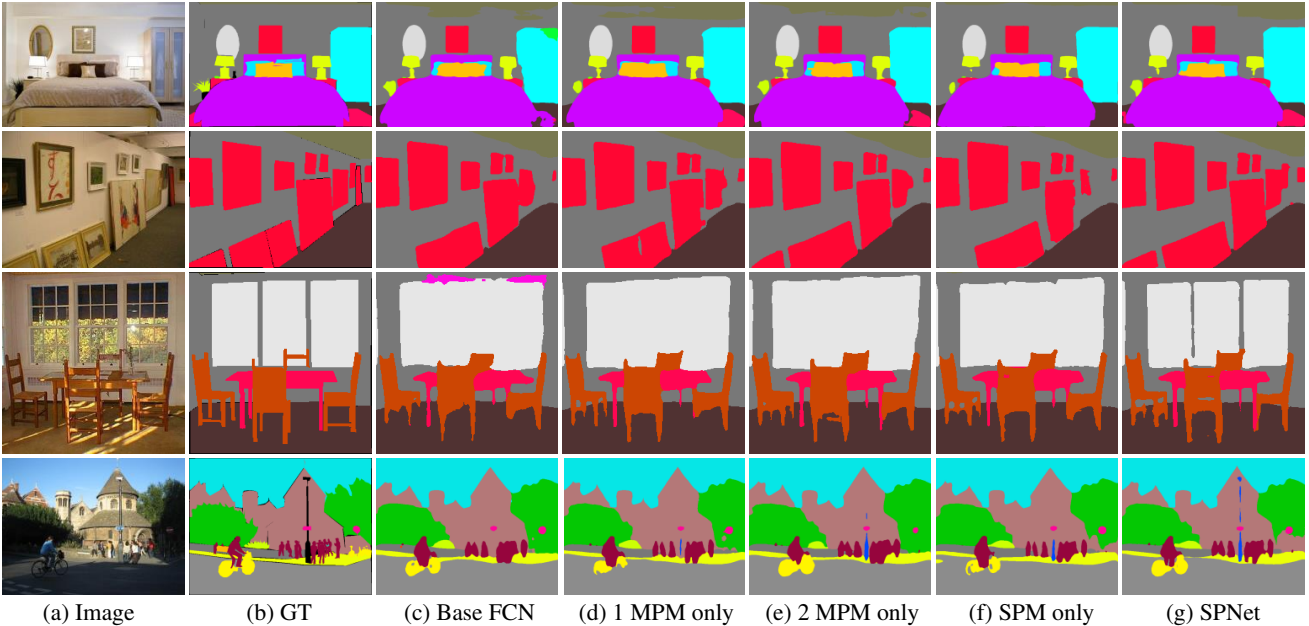

\newparam{Visualization:}
In Figure~\ref{fig:vis_res}, we show some visual results under
different settings of the proposed approach.
Obviously, adding either MPM or SPM to the base FCN can 
effectively improve the segmentation results.
When both MPM and SPM are considered, the quality of the
segmentation maps can be further enhanced.

\begin{table}[tp!]
  \centering
  \small
  \setlength\tabcolsep{1.0mm}
  \renewcommand\arraystretch{1.0}
  \begin{tabular}{lcccc} \toprule[0.7pt]
    Method & Backbone & mIoU (\%) & Pixel Acc. (\%) & Score \\ \midrule[0.5pt]\midrule[0.5pt]
    RefineNet \cite{lin2017refinenet} & ResNet-152 & 40.70 & - & - \\
    PSPNet \cite{zhao2016pyramid} & ResNet-101 & 43.29 & 81.39 & 62.34\\
    PSPNet \cite{zhao2016pyramid} & ResNet-269 & 44.94 & 81.69 & 63.32 \\
    SAC \cite{zhang2017scale} & ResNet-101 & 44.30 & 81.86 & 63.08 \\
    EncNet \cite{zhang2018context} & ResNet-101 & 44.65 & 81.69 & 63.17 \\
    DSSPN \cite{liang2018dynamic} & ResNet-101 & 43.68 & 81.13 & 62.41 \\
    UperNet \cite{xiao2018unified} & ResNet-101 & 42.66 & 81.01 & 61.84 \\
    PSANet \cite{zhao2018psanet} & ResNet-101 & 43.77 & 81.51 & 62.64 \\
    CCNet \cite{huang2018ccnet} & ResNet-101 & 45.22 & - & - \\
    APNB \cite{zhu2019asymmetric} & ResNet-101 & 45.24 & - & - \\
    APCNet \cite{he2019adaptive} & ResNet-101 & 45.38 & - & - \\ \midrule[0.5pt]\midrule[0.5pt]
    SPNet (Ours) & ResNet-50 & 45.03 & 81.32 & 63.18 \\
    SPNet (Ours) & ResNet-101 & \textbf{45.60} & \textbf{82.09} & \textbf{63.85} \\
    \bottomrule[0.7pt]
  \end{tabular}
  \vspace{0pt}
  \caption{Comparisons with the state-of-the-arts on the 
  validation set of ADE20K \cite{zhou2017scene}. We report both mIoU and Pixel Acc. on this benchmark. Best results are highlighted in \textbf{bold}.}
  \label{tab:res_ade20k}
\end{table}

\subsubsection{Comparison with the State-of-the-Arts}

Here, we compare the proposed approach with previous state-of-the-art methods.
The results can be found in Table~\ref{tab:res_ade20k}.
As can be seen, our approach with ResNet-50 as backbone reaches an mIoU score of 45.03\% and pixel accuracy of 81.32\%,
which are already better than most of the previous methods.
When taking ResNet-101 as our backbone, we achieve new state-of-the-art
results in terms of both mIoU and pixel accuracy.

\subsection{Cityscapes}

Cityscapes \cite{cordts2016cityscapes} is another popular dataset 
for scene parsing, which contains totally 19 classes.
It consists of 5K high-quality pixel-annotated images collected from 50 cities in different seasons, all of which are with $1024\times2048$ pixels.
As suggested by previous work, we split the whole dataset into three splits for
training, validation, and test, which contain 2,975, 500, and 1,525 images, respectively.

For a fair comparison, we adopt ResNet-101 as the backbone network.
We compare our approach with existing methods on the test set.
Following previous work \cite{fu2019dual}, we train our network with only fine annotated data and submit the results to the official server.
The results can be found in Table~\ref{tab:test_city}.
It is obvious that the proposed approach outperforms all other methods.

\begin{table}[t]
  \centering
  \small
  \setlength\tabcolsep{1.9mm}
  \renewcommand\arraystretch{1.0}
  \begin{tabular}{lccc} \toprule[0.7pt]
    Method & Publication & Backbone & Test mIoU \\ \midrule[0.5pt]\midrule[0.5pt]
    SAC \cite{zhang2017scale} & ICCV'17 & ResNet-101 & 78.1\% \\
    DUC-HDC \cite{wang2018understanding} & WACV'18 & ResNet-101 & 80.1\% \\
    DSSPN \cite{liang2018dynamic} & CVPR'18 & ResNet-101 & 77.8\% \\
    DepthSeg \cite{kong2018recurrent} & CVPR'18 & ResNet-101 & 78.2\% \\
    DFN \cite{yu2018learning} & CVPR'18 & ResNet-101 & 79.3\% \\
    DenseASPP \cite{yang2018denseaspp}  & CVPR'18 & DenseNet-161 & 80.6\%  \\
    BiSeNet \cite{yu2018bisenet}  & ECCV'18 & ResNet-101 & 78.9\%  \\
    PSANet \cite{zhao2018psanet}  & ECCV'18 & ResNet-101 & 80.1\%  \\
    DANet \cite{fu2019dual}  & CVPR'19 & ResNet-101 & 81.5\%  \\
    SPGNet \cite{cheng2019spgnet} & ICCV'19 & ResNet-101 & 81.1\% \\
    APNB \cite{zhu2019asymmetric} & ICCV'19 & ResNet-101 & 81.3\% \\
    CCNet \cite{huang2018ccnet} & ICCV'19 & ResNet-101 & 81.4\% \\ \midrule[0.5pt]\midrule[0.5pt]
    SPNet (Ours) & - & ResNet-101 & 82.0\% \\
    \bottomrule[0.7pt]
  \end{tabular}
  \vspace{0pt}
  \caption{Comparisons with the state-of-the-arts on the Cityscapes test set \cite{cordts2016cityscapes}.}
  \label{tab:test_city}
  \vspace{-10pt}
\end{table}

\subsection{Pascal Context}

Pascal Context dataset \cite{mottaghi2014role} has 59 categories and 10,103 images with dense label annotations,
which are divided to 4,998 images for training and 5,015 for testing.
Quantitative results can be found in Table~\ref{tab:res_context}.
As can be seen, our approach works much better than other methods.

\begin{table}[t]
  \centering
  \small
  \setlength\tabcolsep{2.5mm}
  \begin{tabular}{cccc} \toprule[0.7pt]
    Method & Publication & Backbone & mIoU (\%) \\ \midrule[0.5pt] \midrule[0.5pt]
    CRF-RNN \cite{zheng2015conditional} & ICCV'15 & VGGNet & 39.3 \\
    BoxSup \cite{dai2015boxsup}  & ICCV'15 & VGGNet & 40.5 \\
    Piecewise \cite{lin2016efficient}  & CVPR'16 & VGGNet & 43.3 \\
    DeepLab-v2 \cite{chen2017deeplab} & PAMI'17 & ResNet-101 & 45.7 \\
    RefineNet \cite{lin2017refinenet} & CVPR'17 & ResNet-152 & 47.3 \\
    CCL \cite{zhang2018context}  & CVPR'18 & ResNet-101& 51.6  \\
    EncNet \cite{zhang2018context}  & CVPR'18 & ResNet-101 & 52.6  \\
    DANet \cite{fu2019dual} & CVPR'19 & ResNet-101 & 52.6 \\
    SVCNet \cite{ding2019semantic} & CVPR'19 & ResNet-101 & 53.2 \\
    EMANet \cite{li2019expectation}  & ICCV'19 & ResNet-101 & 53.1  \\
    APNB \cite{zhu2019asymmetric}  & ICCV'19 & ResNet-101 & 52.8  \\
    BFP \cite{ding2019boundary}  & ICCV'19 & ResNet-101 & 53.6  \\ \midrule[0.5pt] \midrule[0.5pt]
    SPNet (Ours) & - & ResNet-101 & 54.5 \\
    \bottomrule[0.7pt]
  \end{tabular}
  \vspace{0pt}
  \caption{Comparisons with the state-of-the-arts on the Pascal Context dataset \cite{mottaghi2014role}.}
  \label{tab:res_context}
\end{table}

\section{Conclusions}

In this paper, we present a new type of spatial pooling operation, 
strip pooling.
Its long but narrow pooling window allows the model to collect rich global contextual information that is essential for scene parsing networks.
Based on both strip and spatial pooling operations, 
we design a novel strip pooling module to increase the receptive fields
of the backbone network and present a mixed pooling module based on 
the classic residual block with bottleneck structure.
Experiments on several widely-used datasets demonstrate the effectiveness
of the proposed approach.

\myPara{Acknowledgement}
This research was partially supported by AI.SG R-263-000-D97-490, 
NUS ECRA R-263-000-C87-133, MOE Tier-II R-263-000-D17-112,
NSFC (61922046), the national youth talent support program, 
and Tianjin Natural Science Foundation (17JCJQJC43700).

{\small
\bibliographystyle{ieee_fullname}
\bibliography{egbib}
}

\end{document}